\documentclass[letterpaper]{article}
\usepackage[preprint]{aaai2027}
\usepackage[hyphens]{url}
\usepackage{graphicx}
\usepackage{natbib}
\usepackage{caption}
\usepackage{booktabs}
\usepackage{multirow}
\usepackage{amsmath}
\usepackage{amssymb}
\title{Evolving Error States: Failure-Aware Progressive Repair for Ultrasound Lesion Segmentation}
\author{Ziliang Wang\textsuperscript{1}, XuJiang Tang\textsuperscript{2},
Lu Yuting\textsuperscript{3}, Weixin Xu\textsuperscript{3},\\
Yongqiang Zhao\textsuperscript{4}, Ying Fu\textsuperscript{5},
Kehua Guo\textsuperscript{1}\corresponding}
\affiliations{\textsuperscript{1}Central South University,
\textsuperscript{2}Yangtze University,
\textsuperscript{3}Chongqing University,\\
\textsuperscript{4}Peking University,
\textsuperscript{5}Southwest Jiaotong University}

\begin{document}

\maketitle

\newcommand{\substd}[1]{%
  \rlap{\smash{\raisebox{-0.62ex}{\hspace{-0.30em}\scalebox{0.70}{$\scriptscriptstyle\pm#1$}}}}%
}

\begin{abstract}
Reliability under sparse and heterogeneous failures remains a fundamental challenge for medical image segmentation. High average accuracy can conceal a small set of structurally distinct and clinically consequential errors. Existing post-hoc correction methods alleviate this problem, but typically estimate false-positive and false-negative corrections from the same fixed prediction. This ignores the dynamic evolution of error states and limits the correction of complex cases. Inspired by iterative error feedback in structured prediction, we propose \emph{Failure-Aware Progressive Repair} (FAPR). FAPR represents the current segmentation mask as a dynamic failure state and models each repair operation as a state-transition operator. Each accepted correction forms a new prediction state for subsequent error diagnosis and repair, enabling later operations to adapt to preceding changes. Conditional routing selectively activates necessary state transitions, while failure replay exposes the model to rare error states. By keeping the base segmentor frozen, FAPR preserves its established segmentation capability while improving difficult cases. Across three public ultrasound lesion segmentation benchmarks, FAPR improves mean DSC by \textbf{1.52\%}. On the very-hard subsets of BUSI and TN3K, the average gain reaches \textbf{13.77\%}.
\end{abstract}

\section{Introduction}

Encoder--decoder networks, represented by U-Net~\cite{ronneberger2015u} and U-KAN~\cite{li2025u}, have achieved strong average performance in medical image segmentation through multi-scale feature extraction and cross-level feature fusion. However, when applied to ultrasound lesions with weak boundaries, heterogeneous appearance, and substantial shape variation, these models remain susceptible to gross lesion misses, regional under-segmentation, and false-positive predictions. The long-tailed distribution of severe errors further limits their exposure during conventional training. Consequently, high average segmentation accuracy does not necessarily ensure reliable and stable predictions for difficult cases in high-risk clinical settings.

U-Net-style architectures remain a prevalent design paradigm for medical segmentation~\cite{zhou2018unet++,luo2025rethinking}, and recent models such as U-KAN have demonstrated competitive segmentation capability. Most subsequent studies pursue further gains by adding specialized components or introducing increasingly elaborate network designs, as exemplified by MSAGHNet. However, this architecture-centered line of research is primarily evaluated through aggregate metrics and rarely examines whether the same difficult samples remain unresolved. As a result, improvements in mean DSC can coexist with a persistent tail of gross misses, incomplete lesion coverage, and false-positive errors.

\begin{figure}
\centering
\includegraphics[width=\columnwidth]{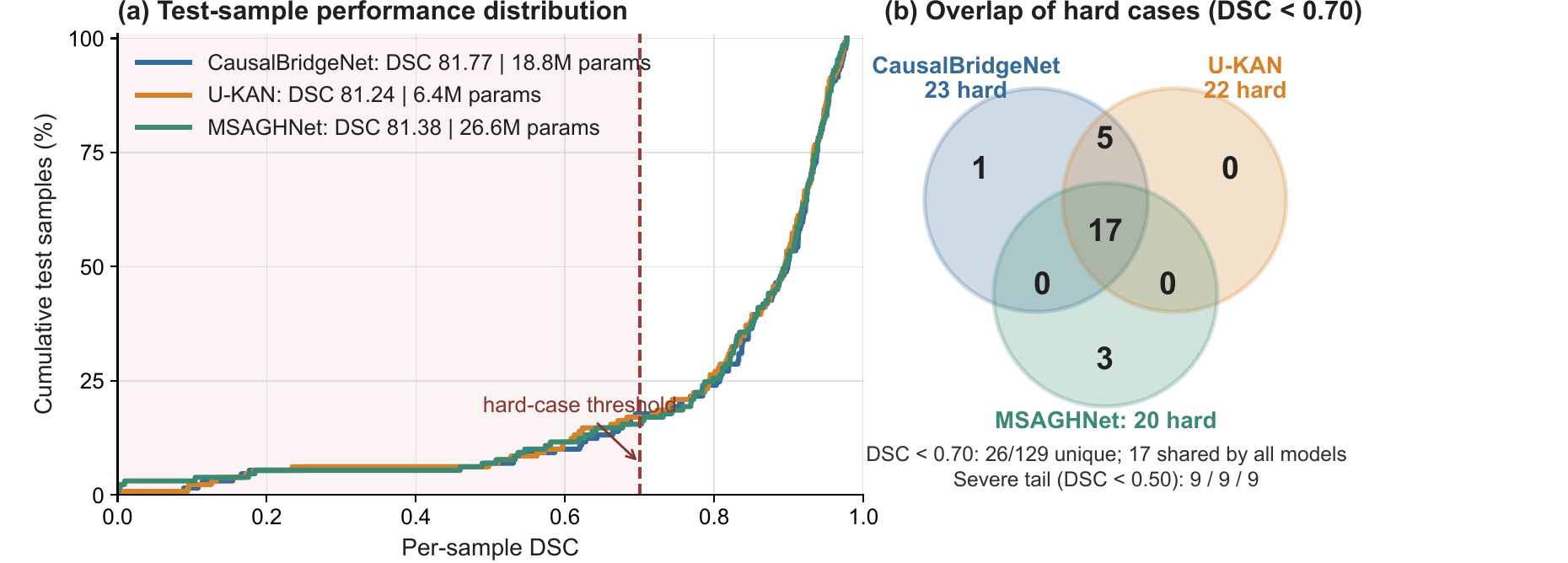}
\caption{Persistent hard-case tails on BUSI. (a) Per-sample DSC distributions with model averages and parameter counts. (b) Among cases below 0.70 DSC, 17 are shared by CausalBridgeNet, U-KAN, and MSAGHNet.}
\label{fig:hard-tail}
\end{figure}

Recent studies have therefore introduced post-hoc correction to repair residual errors without redesigning the base segmentor. CausalBridgeNet~\cite{yang2026make}, for example, estimates false-positive and false-negative corrections from the initial prediction. However, these signals are inferred in parallel from the same fixed state, ignoring how one correction changes the error context for the next. Figure~\ref{fig:hard-tail} summarizes this limitation on BUSI: despite stronger representations in U-KAN and MSAGHNet and the additional post-hoc correction in CausalBridgeNet, their low-accuracy tails remain highly overlapping, with 17 cases below 0.70 DSC for all three methods. Thus, neither architectural scaling nor static correction adequately resolves recurrent structured failures. Once a grossly missed region is recovered, subsequent completion and suppression must instead adapt to the updated spatial support.

Inspired by iterative error feedback~\cite{carreira2016human}, we reinterpret post-hoc correction as an ordered composition of state-transition operators. Unlike conventional mixture-of-experts systems, in which exchangeable experts independently process the same input and compete for selection~\cite{jacobs1991adaptive,riquelme2021scaling,wang2025region}, FAPR assigns heterogeneous experts complementary failure semantics. Each expert diagnoses the latest accepted mask and transforms it into a new prediction state. The repair operators are therefore order-sensitive: recovering a gross miss changes the spatial support available for false-negative completion, which subsequently changes the regions considered by false-positive suppression.

Based on this formulation, we propose Failure-Aware Progressive Repair (FAPR). It decomposes residual errors into gross misses, false-negative incompletion, and false-positive contamination, and organizes the corresponding specialists into an ordered repair chain. Sample- and pixel-level gates control whether and where each transition is accepted, while failure replay provides targeted supervision for rare error states. This design changes expert collaboration from competitive selection to compositional state transition over an evolving segmentation mask.

Our contributions are threefold:
\begin{itemize}
    \item We introduce a state-dependent formulation of medical segmentation repair, modeling correction as progressive transitions between error states rather than a one-shot update of a fixed prediction. It formalizes how each correction changes the spatial support and error context of subsequent operations, providing a principled basis for progressive repair.
    \item We instantiate this formulation with three non-exchangeable specialists for gross-miss recovery, false-negative completion, and false-positive suppression. Dual-level routing controls each transition, while failure replay enables the specialists to learn rare and heterogeneous failure states.
    \item We validate FAPR across three ultrasound lesion benchmarks and different frozen segmentation backbones. Matched controls involving parallel fusion, shared refiners, static-state conditioning, and reversed repair order demonstrate that the gains arise from failure-specific specialization and evolving-state composition rather than additional model capacity.
\end{itemize}

\section{Related Work}

\paragraph{Medical lesion segmentation.}
Medical lesion segmentation has progressed from convolutional encoder--decoder networks toward architectures that incorporate attention, global context, and multi-scale feature interaction~\cite{hatamizadeh2021swin,hatamizadeh2022unetr,li2024deep,miao2023caussl,xu2025scrnet,lu2025layerwise}. U-Net established a symmetric encoder--decoder architecture with skip connections for recovering spatial details~\cite{ronneberger2015u} (\emph{U-Net}). Attention U-Net introduced attention gates to emphasize task-relevant regions during feature fusion~\cite{oktay2018attention} (\emph{Attention U-Net}). TransUNet combined convolutional representations with Transformer-based global context modeling~\cite{chen2021transunet} (\emph{TransUNet}). More recently, MSAGHNet integrated multi-resolution CNN--Transformer representations with scale-guided attention to strengthen feature interaction across spatial scales~\cite{zhu2025multi} (\emph{MSAGHNet}). Although these architectures progressively improve representation learning and average segmentation accuracy, they remain single-pass predictors and do not explicitly revisit heterogeneous residual failures after the initial prediction.

\paragraph{Post-hoc segmentation correction.}
Post-hoc refinement improves an existing segmentation without redesigning its primary prediction pathway. SegFix corrects unreliable boundary pixels using learned directions toward more reliable interior regions~\cite{yuan2020segfix} (\emph{SegFix}). SegRefiner treats coarse-mask refinement as a generative process and progressively updates pixel labels through discrete diffusion~\cite{wang2023segrefiner} (\emph{SegRefiner}). In medical imaging, HiDiff combines a discriminative segmentor with a binary Bernoulli diffusion refiner conditioned on its mask prior~\cite{chen2024hidiff} (\emph{HiDiff}). CausalBridgeNet freezes the foundation model and estimates structured false-positive and false-negative correction signals through a predictive causal reasoning unit~\cite{yang2026make} (\emph{CausalBridgeNet}). These studies establish the value of correcting an initial prediction, but they either target a local error type, employ a homogeneous generic refiner, or derive multiple correction signals from the same prediction prior. They do not explicitly formulate post-hoc correction as ordered transitions among heterogeneous failure states.

\paragraph{Iterative error feedback.}
Iterative Error Feedback introduced a self-correcting formulation in which predicted errors are fed back to progressively update an initial structured output~\cite{carreira2016human} (\emph{Iterative Error Feedback}). Recurrent Mask Refinement subsequently demonstrated in few-shot medical segmentation that the mask produced at one iteration can be reused to recapture changed foreground--background context at the next iteration~\cite{tang2021recurrent} (\emph{Recurrent Mask Refinement}). FAPR builds on this state-updating principle but applies it to automatic post-hoc correction with a frozen segmentor. Instead of repeatedly invoking one shared refiner, it assigns gross-miss recovery, false-negative completion, and false-positive suppression to ordered failure-specific experts, so each repair operates on the state produced by its predecessor.
\begin{figure*}[!t]
\centering
\includegraphics[width=\textwidth]{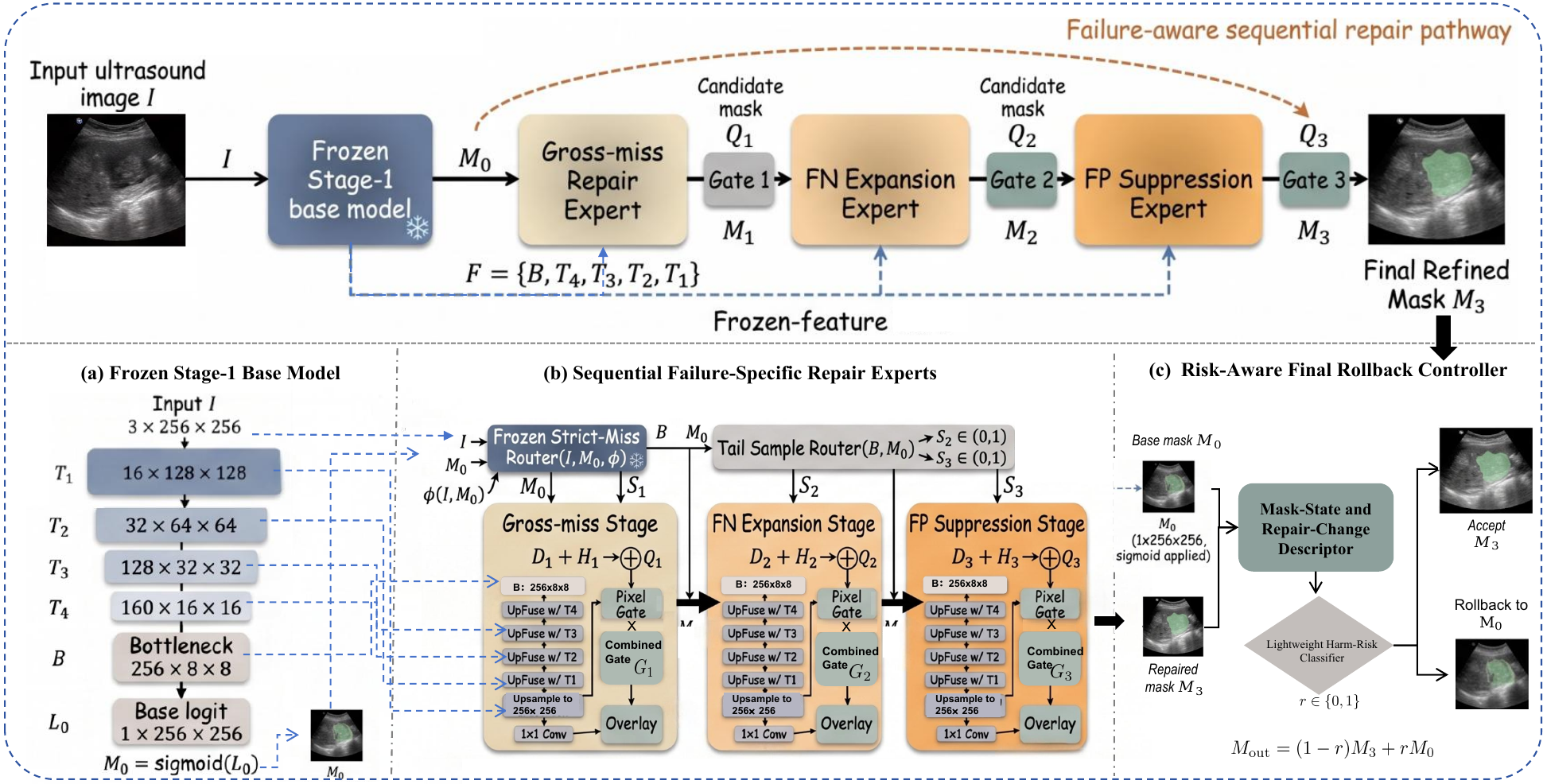}
\caption{Overview of FAPR. The frozen base model provides the initial mask $M_0$ and a shared multi-scale feature pyramid. Three failure-specific experts sequentially propose complete candidate masks $Q_k$, and the corresponding gates selectively overlay each candidate on the latest state to produce $M_1$, $M_2$, and $M_3$. A lightweight risk-aware controller finally accepts $M_3$ or rolls the prediction back to $M_0$.}
\label{fig:overview}
\end{figure*}
\section{Method}

\subsection{Overview}

We propose Failure-Aware Progressive Repair (FAPR), a modular post-hoc framework for correcting a frozen medical image segmentor. Unlike one-shot methods that infer all corrections from the initial prediction, FAPR represents the mask as an evolving failure state and updates it through an ordered pathway of gross-miss recovery, false-negative completion, and false-positive suppression.

As shown in Figure~\ref{fig:overview}, Stage 1 produces the initial mask $M_0$ and frozen multi-scale features. At stage $k$, a failure-specific expert predicts a candidate $Q_k$ from the latest state $M_{k-1}$ and the shared features; sample- and pixel-level routing determine whether and where it is incorporated to form $M_k$. A lightweight final controller rolls harmful trajectories back to $M_0$. Failure replay based on observed Stage-1 errors and controlled perturbations exposes the specialists to rare failures without modifying the base segmentor.

\subsection{Backbone-Agnostic Frozen Base Segmentor}

FAPR is designed as a post-hoc repair framework rather than an architecture-specific extension. It can be attached to different pretrained segmentation models without modifying their original prediction pathways. Given an input image $I\in\mathbb{R}^{3\times H\times W}$, a base segmentor $f_{\theta}$ produces the initial segmentation logits $L_0$ and a hierarchy of intermediate visual features:
\begin{equation}
    (L_0,\mathcal{F})=f_{\theta}(I), \qquad M_0=\sigma(L_0),
\end{equation}
where $M_0\in[0,1]^{H\times W}$ is the initial probability mask and
\begin{equation}
    \mathcal{F}=\{B,T_L,\ldots,T_2,T_1\}
\end{equation}
denotes the feature hierarchy. Here, $B$ represents the bottleneck feature and $T_l$ represents a feature map at the $l$-th spatial resolution. Once Stage 1 is completed, all parameters $\theta$ are frozen, and no gradient from the repair objective is propagated into the base segmentor:
\begin{equation}
    \nabla_{\theta}\mathcal{L}_{\mathrm{repair}}=0.
\end{equation}

The only interface required by FAPR is therefore the initial mask $M_0$ and a set of hierarchical image features $\mathcal{F}$. For encoder-decoder architectures such as U-Net and U-KAN, these features can be directly obtained from their corresponding resolution levels. The repair formulation does not depend on the internal operations used to construct them. Consequently, replacing the base segmentor only changes the feature dimensions exposed through $\mathcal{F}$, while the subsequent failure diagnosis, sequential repair, and state-update process remain unchanged.

In our U-KAN implementation with a $256\times256$ input, the extracted hierarchy is
\begin{equation}
\begin{aligned}
T_1&\in\mathbb{R}^{16\times128\times128}, &
T_2&\in\mathbb{R}^{32\times64\times64},\\
T_3&\in\mathbb{R}^{128\times32\times32}, &
T_4&\in\mathbb{R}^{160\times16\times16},\\
B&\in\mathbb{R}^{256\times8\times8}. &&
\end{aligned}
\end{equation}
The frozen outputs $(M_0,\mathcal{F})$ initialize the repair pathway. At repair stage $k$, the corresponding expert receives the latest prediction state $M_{k-1}$ together with the shared feature hierarchy $\mathcal{F}$, rather than repeatedly invoking or updating the base model.

\subsection{Failure-State Context}

Each specialist receives the frozen feature pyramid and a context representation derived from the current mask. For a probability map $M$, we construct
\begin{equation}
    C(M)=\operatorname{Concat}(M,\mathcal{B}(M),\mathcal{U}(M),\mathcal{A}(M)),
\end{equation}
where $\mathcal{B}$ is a morphological boundary band, $\mathcal{U}(M)=1-2|M-0.5|$ is pixel uncertainty, and $\mathcal{A}(M)$ broadcasts the spatial mean of $M$ as an area map. The context makes the repair conditional on the current prediction geometry rather than only on image features.

\subsection{Sequential Repair Specialists}

At stage $k$, specialist $\mathcal{E}_k$ predicts a complete candidate mask
\begin{equation}
    Q_k=\sigma\left(\mathcal{E}_k(\mathcal{F},C(M_{k-1}))\right).
\end{equation}
All specialists use a five-level decoder with bilinear upsampling, concatenation with the corresponding frozen skip feature, and two Conv-GroupNorm-GELU blocks per level. This preserves the complete candidate-mask capacity that is necessary for large corrections while assigning each specialist a distinct failure objective.

\paragraph{Strict-miss routing.}
Before repair, a separately trained and frozen strict-miss router determines whether $M_0$ lacks reliable lesion support. It jointly analyzes the input image, the base mask, and explicit mask--image statistics $\phi(I,M_0)$:
\begin{equation}
s_1=\sigma\!\left(\mathcal{R}_{\mathrm{miss}}(I,M_0,\phi(I,M_0))\right),
\qquad
S_1=\mathbb{I}[s_1\geq\tau_{\mathrm{miss}}],
\end{equation}
where $S_1\in\{0,1\}$ is a hard sample-level decision. Thus, ordinary predictions bypass global relocalization, while only detected gross misses activate the first repair stage.

\paragraph{Tail-sample routing.}
Unlike gross misses, under-segmentation and over-segmentation usually exhibit gradual rather than binary failure patterns. We therefore employ a shared tail-sample router with stage-specific outputs. Each output is recomputed from the frozen bottleneck feature $B$ and the latest prediction state:
\begin{equation}
\begin{aligned}
\mathbf{u}_{k-1}
&=\left[\operatorname{GAP}(B),\operatorname{GMP}(B),\psi(M_{k-1})\right],\\
S_k&=\sigma\!\left(\mathcal{R}_{\mathrm{tail},k}(\mathbf{u}_{k-1})\right),
\qquad k\in\{2,3\}.
\end{aligned}
\end{equation}
Here, $\operatorname{GAP}$ and $\operatorname{GMP}$ denote global average and max pooling, respectively, and $\psi(M_{k-1})$ contains confidence, foreground-area, and uncertainty statistics of the current mask. Thus, $S_2=\mathcal{R}_{\mathrm{tail},2}(B,M_1)$ diagnoses the state after gross-miss repair, whereas $S_3=\mathcal{R}_{\mathrm{tail},3}(B,M_2)$ diagnoses the state after FN expansion. The soft scores $S_2,S_3\in(0,1)$ control the sample-level activation strengths of the corresponding stages, while their pixel-level gates determine the spatial extent of each repair.

\paragraph{Gross-miss repair stage.}
When $S_1=1$, the gross-miss expert reconstructs a plausible lesion candidate from the frozen multi-scale features and the current mask context:
\begin{equation}
Q_1=\sigma\!\left(
\mathcal{D}_{\mathrm{miss}}(\mathcal{F})
+\mathcal{H}_{\mathrm{miss}}(C(M_0))
\right),
\end{equation}
where $\mathcal{D}_{\mathrm{miss}}$ reconstructs global lesion structure and $\mathcal{H}_{\mathrm{miss}}$ encodes the geometry and uncertainty of $M_0$. A pixel gate $P_1$ selects the spatial support of the candidate, and the combined gate $G_1=S_1\odot P_1$ updates the state as
\begin{equation}
M_1=(1-G_1)\odot M_0+G_1\odot Q_1.
\end{equation}
The resulting $M_1$ is then passed to the subsequent refinement stages.

\paragraph{False-negative expansion stage.}
The second expert operates on the updated state $M_1$ and recovers lesion regions that remain under-segmented after gross-miss repair. Rather than estimating a correction from the obsolete base mask, it predicts a complete candidate $Q_2$ from the shared frozen features and the context of $M_1$:
\begin{equation}
Q_2=\sigma\!\left(
\mathcal{D}_{\mathrm{fn}}(\mathcal{F})
+\mathcal{H}_{\mathrm{fn}}(C(M_1))
\right).
\end{equation}
The pixel gate evaluates the proposed change against the latest state:
\begin{equation}
P_2=\sigma\!\left(
\mathcal{G}_{\mathrm{fn}}
\left([M_1,Q_2,Q_2-M_1,(1-M_1)\odot Q_2]\right)
\right).
\end{equation}
It is combined with the tail-sample score $S_2$ to obtain $G_2=S_2\odot P_2$. The next prediction state is then
\begin{equation}
M_2=(1-G_2)\odot M_1+G_2\odot Q_2.
\end{equation}
Consequently, expansion is activated only for samples exhibiting an under-segmentation tendency and only at spatial locations supported by the current candidate. The updated mask $M_2$ is passed to the false-positive suppression stage.

\paragraph{False-positive suppression stage.}
The final expert removes unsupported responses from the latest state $M_2$, including over-segmented regions and isolated false-positive components introduced by either the base model or preceding repairs. It predicts a complete suppression candidate from the shared frozen features and the context of $M_2$:
\begin{equation}
Q_3=\sigma\!\left(
\mathcal{D}_{\mathrm{fp}}(\mathcal{F})
+\mathcal{H}_{\mathrm{fp}}(C(M_2))
\right).
\end{equation}
The corresponding pixel gate compares the candidate with the current prediction state:
\begin{equation}
P_3=\sigma\!\left(
\mathcal{G}_{\mathrm{fp}}
\left([M_2,Q_3,Q_3-M_2,M_2\odot(1-Q_3)]\right)
\right).
\end{equation}
Here, $M_2\odot(1-Q_3)$ explicitly represents the foreground region in the current state that the candidate proposes to remove. Combining the pixel gate with the tail-sample score gives $G_3=S_3\odot P_3$, and the final repaired state is
\begin{equation}
M_3=(1-G_3)\odot M_2+G_3\odot Q_3.
\end{equation}
This final transition suppresses unsupported foreground responses only where both sample-level diagnosis and pixel-level evidence permit modification, while retaining the preceding state elsewhere.

\subsection{Risk-Aware Rollback}

Sequential repair may occasionally perturb an already reliable base mask. We therefore introduce a lightweight controller that decides whether to accept $M_3$ or restore $M_0$. It uses the base-mask state and the accumulated repair change:
\begin{equation}
\mathbf{z}=[\phi(M_0),\Delta(M_0,M_3)],
\qquad
p_{\mathrm{harm}}=\mathcal{R}_{\eta}(\mathbf{z}),
\end{equation}
where $\phi(M_0)$ contains four base-state descriptors: foreground area ratio, vertical centroid, largest-component ratio, and mean predictive entropy. The change descriptor $\Delta(M_0,M_3)$ contains the add--remove balance, area-ratio change, connected-component change, boundary-density change, scale-normalized centroid shift, and low overlap. The controller is supervised by
\begin{equation}
y^{\mathrm{harm}}=\mathbb{I}\!\left[\operatorname{DSC}(M_3,Y)<\operatorname{DSC}(M_0,Y)\right],
\end{equation}
with sample weight $w=\max(|\operatorname{DSC}(M_3,Y)-\operatorname{DSC}(M_0,Y)|,10^{-3})$. Candidate lightweight classifiers are fitted only on deduplicated training predictions. The classifier family and decision threshold are selected only on the validation set by maximizing rollback DSC, subject to a maximum rollback rate of $5\%$. Both are then frozen before test evaluation. At inference,
\begin{equation}
r=\mathbb{I}[p_{\mathrm{harm}}\ge\tau],
\qquad
M_{\mathrm{out}}=(1-r)M_3+rM_0.
\end{equation}
The controller uses no ground-truth information at inference. It makes one final decision after the complete sequential repair and never rolls back intermediate states.

\subsection{Failure Replay and Controlled Mask Corruption}

Severe segmentation failures are sparse in the original training distribution. We therefore construct a training-only failure replay set from predictions of the frozen Stage-1 model. For each training pair $(I_i,Y_i)$, the base segmentor produces $M_i^0$. Let
\begin{equation}
d_i^0=\operatorname{DSC}(M_i^0,Y_i).
\end{equation}
The hard subset is defined as
\begin{equation}
\mathcal{H}=\{(I_i,Y_i)\in\mathcal{D}_{\mathrm{train}}:d_i^0<0.7\}.
\end{equation}
Each selected example is replayed once together with the original training set:
\begin{equation}
\mathcal{D}_{\mathrm{replay}}
=\mathcal{D}_{\mathrm{train}}\cup\mathcal{H}.
\end{equation}
The discrepancy between $M_i^0$ and $Y_i$ supplies false-negative and false-positive supervision for the routers and pixel gates. Because severe failures are sparse, the Stage-1 prior is replaced with probability $p_c=0.55$ by a controlled ground-truth corruption sampled from four modes: gross miss, under-segmentation, over-segmentation, or boundary jitter. These modes are implemented using erosion or cutout, dilation or disconnected blobs, and sparse boundary perturbations. Otherwise, the real Stage-1 prediction is retained. Replay construction uses training annotations only; validation and test samples are excluded, and inference always starts from the frozen segmentor's prediction.

\subsection{Training Objectives}

FAPR jointly supervises the candidate masks, sequential states, and routing modules. Let $Y$ denote the ground-truth mask. Each specialist is optimized using region-weighted binary cross-entropy and Tversky loss:
\begin{equation}
\mathcal{L}_{\mathrm{exp}}
=
\sum_{k=1}^{3}
\left[
\mathcal{L}_{\mathrm{WBCE}}(Q_k,Y;W_k)
+
\mathcal{L}_{\mathrm{Tv}}(Q_k,Y;\alpha_k,\beta_k)
\right].
\end{equation}
Here, $W_k$ emphasizes the failure region assigned to expert $k$. The gross-miss and FN experts use recall-oriented supervision, whereas the FP expert uses precision-oriented supervision. All intermediate states $M_k$ and the final output $M_3$ are additionally supervised by BCE and Dice losses.

The pixel gates are trained against their corresponding gross-miss, FN, and FP regions. Sample routers use class-balanced binary cross-entropy with failure labels derived from Stage-1 training predictions. A preservation term penalizes changes to correctly predicted regions, while sparsity regularization discourages unnecessary repair. The complete objective is
\begin{equation}
\begin{aligned}
\mathcal{L}={}&
\lambda_e\mathcal{L}_{\mathrm{exp}}
+\lambda_s\sum_{k=1}^{3}\mathcal{L}_{\mathrm{seg}}(M_k,Y)
+\lambda_f\mathcal{L}_{\mathrm{seg}}(M_3,Y)\\
&+\lambda_g\mathcal{L}_{\mathrm{gate}}
+\lambda_r\mathcal{L}_{\mathrm{router}}
+\lambda_p\mathcal{L}_{\mathrm{pres}}
+\lambda_{\mathrm{sp}}\mathcal{L}_{\mathrm{sparse}},
\end{aligned}
\end{equation}
where $\mathcal{L}_{\mathrm{seg}}=\mathcal{L}_{\mathrm{BCE}}+\mathcal{L}_{\mathrm{Dice}}$. The strict-miss router is pretrained and frozen before specialist training. The risk-aware rollback controller is calibrated separately after sequential repair training and does not propagate gradients to the base segmentor or repair modules.
\begin{table*}
\centering
{
\scriptsize
\setlength{\tabcolsep}{1.2pt}
\begin{tabular*}{\textwidth}{@{\extracolsep{\fill}}lccccccccccccc@{}}
\toprule
& & \multicolumn{4}{c}{BUSI} & \multicolumn{4}{c}{BUSIS} & \multicolumn{4}{c}{TN3K} \\
\cmidrule(lr){3-6}\cmidrule(lr){7-10}\cmidrule(lr){11-14}
Method & Params (M) & DSC & mIoU & Prec. & Rec. & DSC & mIoU & Prec. & Rec. & DSC & mIoU & Prec. & Rec. \\
\midrule
U-Net (Ronneberger et al., 2015) & 31.04 & 80.91 & 72.16 & 81.84 & 84.77 & 91.18 & 84.89 & 93.02 & 90.88 & 80.48 & 71.04 & 82.26 & 84.53 \\
AttUNet (Oktay et al., 2018) & 34.88 & 76.62 & 68.09 & 79.72 & 78.44 & 91.04 & 84.65 & 93.04 & 90.66 & 77.80 & 67.86 & 74.94 & 87.04 \\
SegRefiner (Wang et al., 2023) & 124.96 & 81.34 & 72.46 & 83.35 & 83.71 & 92.63 & 86.80 & 93.33 & 92.76 & 81.91 & 73.23 & 85.87 & 84.38 \\
CMUNet (Tang et al., 2023) & 49.93 & 79.95 & 71.68 & 84.13 & 81.45 & 91.43 & 85.28 & 92.86 & 91.58 & 79.86 & 70.15 & 78.35 & 85.19 \\
MSAGHNet (Zhu et al., 2025) & 26.61 & 81.44 & 72.53 & 84.04 & 81.78 & 91.36 & 84.76 & 92.88 & 90.94 & 81.01 & 71.35 & 79.83 & 87.55 \\
U-KAN (Li et al., 2025) & 6.36 & 81.24 & 72.29 & 83.50 & 83.35 & 92.50 & 86.64 & 93.25 & 92.65 & 81.83 & 73.10 & 85.37 & 84.48 \\
CausalBridgeNet (Yang et al., 2026) & 18.76 & 81.77 & 73.03 & 84.06 & 83.44 & 92.53 & 86.68 & 93.14 & \textbf{92.79} & 82.12 & 73.47 & 85.19 & 85.25 \\
\midrule
FAPR (Ours) & 17.37 & \textbf{83.78}\substd{0.2} & \textbf{75.34}\substd{0.3} & \textbf{86.08}\substd{0.2} & \textbf{84.51}\substd{0.2} & \textbf{92.62}\substd{0.1} & \textbf{86.79}\substd{0.2} & \textbf{93.66}\substd{0.1} & 92.41\substd{0.2} & \textbf{83.72}\substd{0.1} & \textbf{75.15}\substd{0.1} & \textbf{85.29}\substd{0.1} & \textbf{86.67}\substd{0.1} \\
\bottomrule
\end{tabular*}
}
\caption{Comparison on BUSI, BUSIS, and TN3K. Values are percentages and Params denotes the total number of parameters in millions. The FAPR results are averaged over three random seeds; tiny subscripts denote standard deviations.}
\label{tab:main}
\end{table*}

\section{Experiments}

\subsection{Datasets}

\paragraph{BUSI.}
BUSI~\cite{al2020dataset} contains breast ultrasound images with benign, malignant, and normal cases. Following the fixed lesion-only split used by our baselines, we use 453 training, 65 validation, and 129 test images.

\paragraph{TN3K.}
TN3K~\cite{gong2021tn3k} contains 3,493 thyroid nodule ultrasound images. We follow its official test split, using 2,519 training, 360 validation, and 614 test images.

\paragraph{BUSIS.}
BUSIS~\cite{zhang2022busis} is a breast ultrasound segmentation benchmark containing 562 de-identified images collected from multiple hospitals. We use 394 images for training, 56 for validation, and 112 for testing.

\subsection{Baselines}

We compare FAPR with representative segmentation networks and post-hoc refiners. U-Net~\cite{ronneberger2015u}, Attention U-Net~\cite{oktay2018attention}, CMUNet~\cite{tang2023cmunet}, MSAGHNet~\cite{zhu2025multi}, and U-KAN~\cite{li2025u} cover convolutional, attention-based, and KAN-based segmentors, while SegRefiner~\cite{wang2023segrefiner} represents generic mask refinement. CausalBridgeNet~\cite{yang2026make} is the closest recent correction baseline: its PCRU processes the same prompt-conditioned decoder feature through four parallel causal experts~\cite{jacobs1991adaptive,riquelme2021scaling,wang2025region} and mixes their outputs to estimate FP and FN maps for a one-shot correction of the initial logits. In contrast, FAPR decomposes the correction into three ordered failure-specific stages, each conditioned on the latest accepted mask. Our matched CausalBridgeNet reproduction uses the same frozen U-KAN, data splits, input resolution, and validation-based checkpoint selection as FAPR.

\subsection{Evaluation Metrics}

We report DSC, mIoU, Precision, and Recall, computed per image and averaged across the evaluation set. Checkpoints are selected by validation DSC and evaluated once on the held-out test set.

\subsection{Implementation Details}
FAPR is implemented in PyTorch and trained on two NVIDIA RTX PRO 5000 Blackwell GPUs with a global batch size of 16. Table~\ref{tab:main} reports averages over seeds 2981, 3407, and 42. The Stage-1 segmentor and pretrained strict-miss router remain frozen. Repair modules are trained for 80 epochs using Adam with an initial learning rate of $8\times10^{-5}$, weight decay of $10^{-4}$, cosine decay to $10^{-6}$, and gradient clipping at 5. Checkpoints are selected by validation DSC. Replay samples and router labels use training data only, while the final rollback controller is calibrated on the training and validation splits.

Under single-image FP32 inference at $256\times256$ resolution, FAPR uses 7.4\% fewer parameters than the matched CausalBridgeNet reproduction and reduces Conv/Linear FLOPs, latency, and peak memory by 81.3\%, 59.3\%, and 73.9\%, respectively.

Additional implementation details, code, training scripts, and data splits are provided in the supplementary material. Code and trained weights will be released upon acceptance.

\subsection{Effectiveness of FAPR}

Tables~\ref{tab:main} and~\ref{tab:hard} report aggregate and failure-conditioned performance on BUSI and TN3K. Hard and very-hard cases satisfy $\mathrm{DSC}(M_0,Y)<0.7$ and $<0.5$, respectively; these labels are used only for analysis.

\begin{table}
\centering
{
\footnotesize
\begin{tabular*}{\columnwidth}{@{\extracolsep{\fill}}llccc@{}}
\toprule
Dataset & Method & Very hard & Hard & Easy \\
\midrule
\multirow{4}{*}{BUSI}
& Stage 1 & 17.04 & 42.62 & 89.18 \\
& SegRefiner & 17.22 & 42.54 & 89.32 \\
& CausalBridgeNet & 17.20 & 43.80 & 89.58 \\
& FAPR & \textbf{35.78}\substd{0.4} & \textbf{53.69}\substd{1.0} & \textbf{90.11}\substd{0.0} \\
\midrule
\multirow{4}{*}{TN3K}
& Stage 1 & 25.97 & 42.85 & 89.60 \\
& SegRefiner & 25.93 & 42.82 & 89.68 \\
& CausalBridgeNet & 26.99 & 43.69 & \textbf{89.78} \\
& FAPR & \textbf{34.76}\substd{2.7} & \textbf{49.39}\substd{0.8} & 89.25\substd{0.1} \\
\bottomrule
\end{tabular*}
}
\caption{Failure-conditioned DSC. Columns denote very-hard ($<0.5$), hard ($<0.7$), and easy ($\geq0.7$) subsets. Their sizes are BUSI: 9/22/107 and TN3K: 53/102/512.}
\label{tab:hard}
\end{table}

Using the same frozen U-KAN, FAPR outperforms the matched static correctors on both benchmarks (Table~\ref{tab:main}). Its advantage is concentrated in the low-accuracy tail: very-hard DSC increases by 18.74 points on BUSI and 8.79 points on TN3K, whereas easy-case DSC changes by only $+0.93$ and $-0.35$ points, respectively (Table~\ref{tab:hard}). This separation supports progressive repair of structural failures rather than uniform smoothing of all predictions.

\subsection{Robustness of FAPR}

BUSIS is nearly saturated and contains only one hard test case. FAPR reaches 92.62 DSC and 86.79 mIoU, within 0.01 points of SegRefiner and slightly above the frozen Stage-1 model. Thus, selective routing preserves performance when little repair is needed.
\begin{figure*}
\centering
\includegraphics[width=\textwidth]{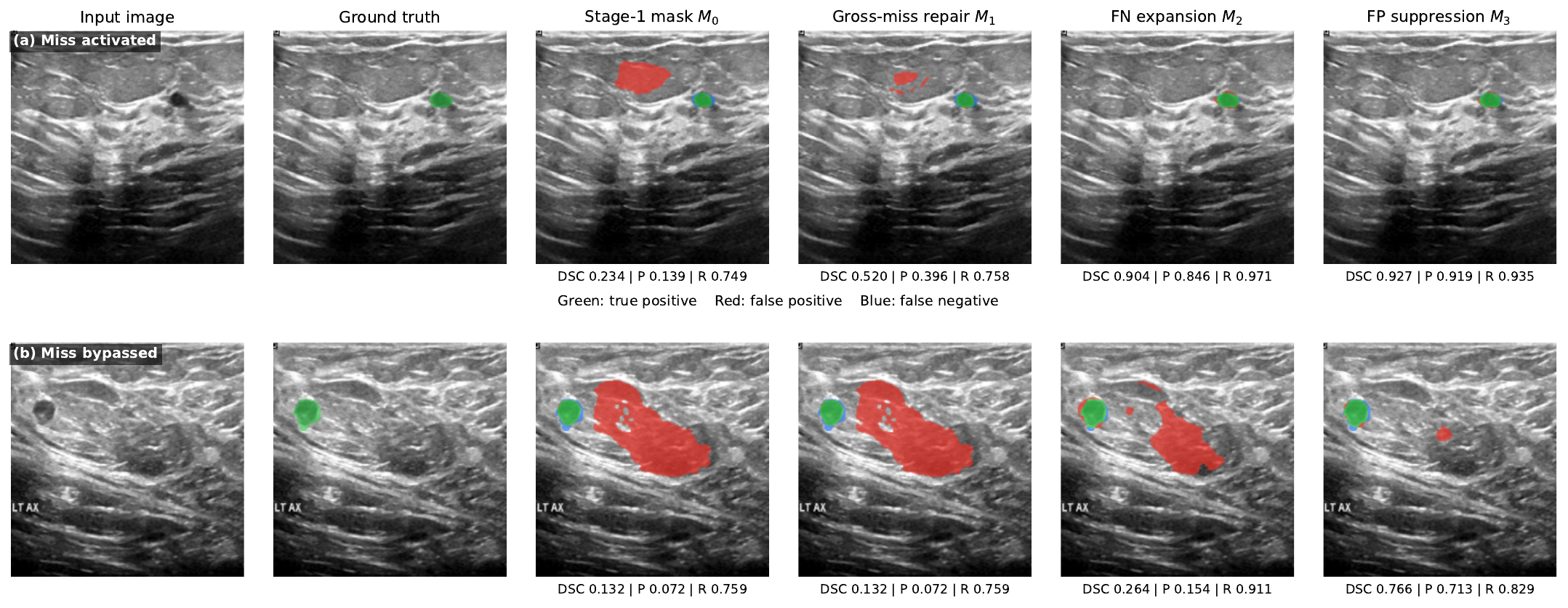}
\caption{Qualitative prediction-state progression on two BUSI test cases. True-positive, false-positive, and false-negative regions are shown in green, red, and blue, respectively. (a) The strict-miss route is activated. (b) The miss route is bypassed, so $M_1=M_0$, while the FN-expansion and FP-suppression specialists perform the subsequent repair.}
\label{fig:qualitative_progression}
\end{figure*}

\subsection{Roles of the Sequential Specialists}

\begin{table}
\centering
{
\footnotesize
\setlength{\tabcolsep}{4pt}
\begin{tabular}{lcccc}
\toprule
Configuration & Overall & Easy & Hard & Very hard \\
\midrule
FAPR & \textbf{83.78}\substd{0.2} & \textbf{90.11}\substd{0.0} & \textbf{53.69}\substd{1.0} & \textbf{35.78}\substd{0.4} \\
w/o Gross-miss & 83.28 & \textbf{90.03} & 50.49 & 30.40 \\
w/o FN Expansion & 82.48 & 89.77 & 47.00 & 21.58 \\
w/o FP Suppression & 82.31 & 89.84 & 45.67 & 21.66 \\
w/o Failure Replay & 82.38 & 89.90 & 45.83 & 18.95 \\
w/o Controlled Corruption & 82.40 & 89.89 & 45.97 & 19.51 \\
\bottomrule
\end{tabular}
}
\caption{Specialist and training ablations on the BUSI test set. The complete FAPR row includes the final rollback controller.}
\label{tab:specialist_ablation}
\end{table}

Table~\ref{tab:specialist_ablation} isolates the contribution of the repair components. Removing individual specialists lowers very-hard DSC by 5.38--14.20 points, while removing replay or controlled corruption lowers it by 16.27--16.83 points. Easy-case DSC remains within 89.77--90.11, showing that these components primarily repair difficult failures.

\noindent\textbf{Sparse gross-miss routing.}
The strict-miss router activates for only 3/129 BUSI test cases, yet all three improve after the first repair stage, with a mean $M_0\!\rightarrow\!M_1$ gain of 9.75 DSC points. Removing the gross-miss specialist further reduces very-hard DSC by 5.38 points. Thus, the miss route is rarely used but remains important for recovering extreme localization failures.

\noindent\textbf{Low-cost rollback.}
The final controller uses only ten mask-change descriptors and requires neither additional image encoding nor another segmentation forward pass. Despite its negligible computational cost, it improves DSC by 0.13 and 0.004 percentage points on BUSI and TN3K, respectively, providing a conservative safety gain after sequential repair. On the nearly saturated BUSIS benchmark, where hard failures are almost absent, no prediction is rolled back.

\begin{table}[!htbp]
\centering
{
\footnotesize
\setlength{\tabcolsep}{1.5pt}
\begin{tabular*}{\columnwidth}{@{\extracolsep{\fill}}lcccccc@{}}
\toprule
Configuration & Total & Train. & DSC & mIoU & Hard & Very hard \\
 & (M) & (M) & & & DSC & DSC \\
\midrule
U-KAN-Large (GB16) & 25.36 & 25.36 & 78.76 & 68.85 & 47.36 & 28.96 \\
Shared Refiner $\times3$ & 17.32 & 10.70 & 82.69 & 74.17 & 48.01 & 27.77 \\
Parallel Expert Fusion & 17.61 & 10.99 & 82.99 & 74.41 & 48.84 & 29.21 \\
Static-$M_0$ Cascade & 17.37 & 10.75 & 82.98 & 74.45 & 50.88 & 33.61 \\
Reverse Repair Order & 17.37 & 10.75 & 82.55 & 73.92 & 46.02 & 21.83 \\
Uniform BCE+Dice & 17.37 & 10.75 & 82.02 & 73.39 & 43.90 & 17.49 \\
\midrule
FAPR & 17.37 & 10.75 & \textbf{83.78} & \textbf{75.34} & \textbf{53.69} & \textbf{35.78} \\
\bottomrule
\end{tabular*}
}
\caption{Mechanism ablation study on the BUSI test set.}
\label{tab:capacity_mechanism_controls}
\end{table}

Table~\ref{tab:capacity_mechanism_controls} separates model capacity from the proposed repair mechanism. A larger backbone, a repeated shared refiner, and parallel expert fusion all underperform FAPR, showing that the gains do not arise from additional capacity alone. Conditioning every specialist on the static $M_0$ weakens hard-case repair, confirming the importance of evolving-state inputs. Reversing the repair order or replacing failure-specific objectives with uniform supervision causes a larger degradation, demonstrating that the specialists are order-sensitive and non-exchangeable. Together, these controls validate the coordinated roles of failure specialization, progressive state updating, and ordered execution.

\noindent\textbf{Visualization.} To better illustrate the repair workflow of FAPR, in Figure~\ref{fig:qualitative_progression}, the first case activates gross-miss recovery before expansion and suppression, while the second bypasses it and is repaired by the later specialists. These trajectories illustrate that FAPR follows sample-dependent repair paths rather than applying every correction indiscriminately.

\begin{table}
\centering
{
\footnotesize
\setlength{\tabcolsep}{2.5pt}
\begin{tabular}{llcccc}
\toprule
Dataset & Method & DSC & mIoU & Hard & Very hard \\
\midrule
\multirow{2}{*}{BUSI}
 & U-Net Stage1 & 80.91 & 72.16 & 38.36 & 16.38 \\
 & U-Net + FAPR & \textbf{82.86} & \textbf{74.71} & \textbf{46.47} & \textbf{30.37} \\
\cmidrule(lr){1-6}
\multirow{2}{*}{TN3K}
 & U-Net Stage1 & 80.48 & 71.04 & 42.38 & 21.83 \\
 & U-Net + FAPR & \textbf{82.55} & \textbf{73.22} & \textbf{52.19} & \textbf{37.83} \\
\bottomrule
\end{tabular}
}
\caption{Backbone-transfer results with a frozen U-Net Stage-1 model.}
\label{tab:unet_transfer}
\end{table}

\noindent\textbf{Backbone transfer.} With a frozen U-Net, FAPR improves overall DSC by 1.95 points on BUSI and 2.07 points on TN3K. The gains increase to 8.11/13.99 points on BUSI and 9.81/16.00 points on the hard/very-hard subsets, confirming that the repair mechanism transfers beyond U-KAN and remains focused on the low-accuracy tail.

\section{Conclusion}

We presented FAPR, a post-hoc framework that models ultrasound segmentation correction as progressive prediction-state transitions. Ordered specialists recover gross misses, complete FN regions, and suppress FP regions, while selective routing limits harmful updates. Experiments on three datasets and two frozen backbones show consistent gains concentrated on difficult cases without sacrificing saturated benchmarks. FAPR thus improves difficult predictions without retraining the base segmentor. These findings suggest that residual segmentation failures are better modeled as coupled, evolving states than as independent corrections derived from a fixed prediction. Future work will extend this formulation to stronger volumetric repair for broader medical imaging applications by explicitly modeling 3D spatial continuity and inter-slice dependencies.

\enlargethispage{2\baselineskip}
\bibliography{references}


\end{document}